\documentclass[10pt,twocolumn,letterpaper]{article}

\usepackage[pagenumbers]{cvpr} 

\usepackage[dvipsnames]{xcolor}
\usepackage{xspace}

\usepackage{dsfont}
\usepackage{tabularx}
\usepackage{multirow}

\newcommand{\R}{\ensuremath{\mathbb{R}}\xspace}
\newcommand{\ud}{\ensuremath{\mathrm{d}}\xspace}
\newcommand{\vx}{\ensuremath{\vec{x}}\xspace}
\newcommand{\vB}{\ensuremath{\vec{B}_t}\xspace}

\definecolor{cvprblue}{rgb}{0.21,0.49,0.74}
\usepackage[pagebackref,breaklinks,colorlinks,citecolor=cvprblue]{hyperref}

\title{On the Noise Scheduling for Generating Plausible Designs with Diffusion Models}

\author{Jiajie Fan$^{1,2}$, \quad Laure Vuaille, \quad Thomas Bäck$^{2}$, \quad Hao Wang$^{2}$\\
\textsuperscript{1} BMW Group\\
\textsuperscript{2} LIACS, Leiden University\\
{\tt\small jiajie.fan@bmw.de, laure.vuaille@tum.de, \{t.h.w.baeck, h.wang\}@liacs.leidenuniv.nl}
}

\begin{document}
\maketitle
\begin{abstract}
Deep Generative Models (DGMs) are widely used to create innovative designs across multiple industries, ranging from fashion to the automotive sector. In addition to generating images of high visual quality, the task of structural design generation imposes more stringent constrains on the semantic expression, \eg, no floating material or missing part, which we refer to as plausibility in this work. We delve into the impact of noise schedules of diffusion models on the plausibility of the outcome: there exists a range of noise levels at which the model's performance decides the result plausibility. Also, we propose two techniques to determine such a range for a given image set and devise a novel parametric noise schedule for better plausibility. We apply this noise schedule to the training and sampling of the well-known diffusion model EDM and compare it to its default noise schedule. Compared to EDM, our schedule significantly improves the rate of plausible designs from 83.4\% to 93.5\% and Fr\'echet Inception Distance (FID) from 7.84 to 4.87. Further applications of advanced image editing tools demonstrate the model's solid understanding of structure.



 
 

\end{abstract}    
\section{Introduction}
\label{sec:intro}
\begin{figure}[t]
\fontsize{8}{10}\selectfont
  \centering
             \begin{subfigure}[b]{\linewidth}
            \centering
            \includegraphics[width=\linewidth]{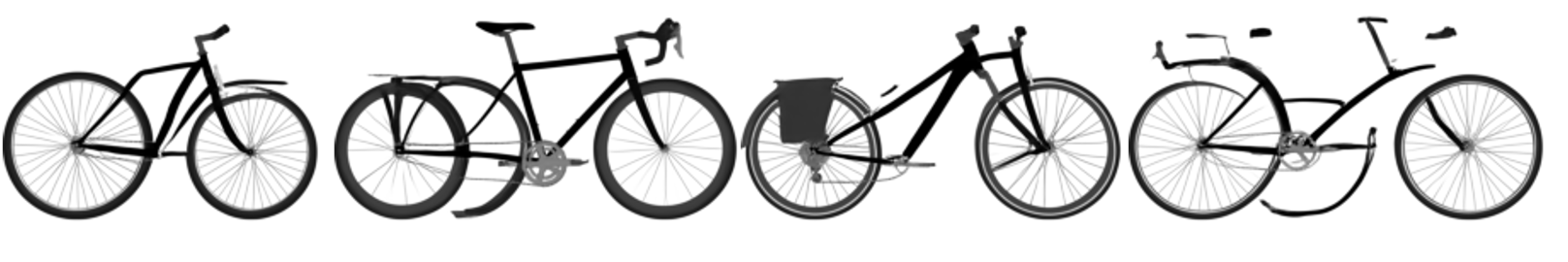}
            \caption{Implausible bicycles.}
            \label{fig:semantic_error_EDM}
            \end{subfigure}\\
            \begin{subfigure}[b]{\linewidth}
            \centering
            \includegraphics[width=\linewidth]{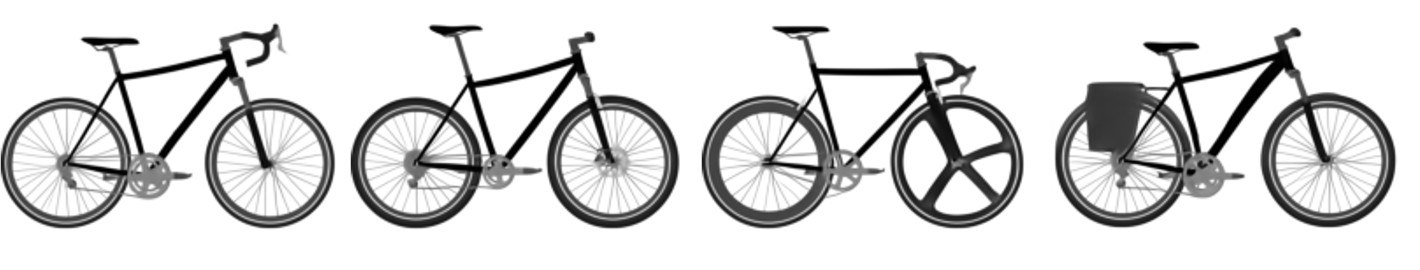}
            \caption{Plausible bicycles.}
            \label{fig:semantic_oriented_PoDM}
            \end{subfigure}
            \caption{Generated bicycle designs. Our work aims to minimize the proportion of implausible designs generated.}
            \label{fig:feasiblity_examples}
  \end{figure}

Deep Generative Models (DGMs) have emerged as a powerful algorithm for enabling Generative Design, hereby modeling and exploring design spaces. This approach has also been employed for structural designs and shown its potential in synthesizing complex structures~\cite{generativeDesign2021Regenwetter, wei2020padgan,wei2018BezierGAN,fan2023adversarial}. In design generation tasks, especially for structures, DGMs may generate implausible designs,~\eg, bicycles with an extra handle, missing wheel or an invalid layout as shown in the first row of~\cref{fig:feasiblity_examples}, which need to be avoided. However, plausibility of generated images has not been sufficiently studied. Instead, we argue that recent DGMs are unnecessarily convoluted in order to obtain better visual quality, which is primarily measured with the metric Fr\'echet Inception Distance (FID)~\cite{heusel2018TTUR}.

Also, we note that in the field of Generative Design, the implemented models tend to be based on Generative Adversarial Networks (GANs)~\citep{GAN} and are therefore suffering from same issues as GANs, \eg, unstable training, mode collapse and low diversity~\citep{GAN,stabilizingGAN,tutorialGAN,spectralNormalization}. Meanwhile, diffusion-based generative models, including a fixed forward process that gradually corrupts source images by adding noise and a trainable generation process that iteratively reverses the perturbation, have surpassed GANs and become the state-of-the-art (SOTA) generative model in various image synthesis tasks~\cite{ho2020ddpm, song2022ddim,dhariwal2021diffusion, karras2022elucidating}. Recent studies~\cite{nichol2021improved, song2020improved, song2021scorebased, karras2022elucidating} have explored the effect of noise level on properties of generated images, \eg, the quality of image details depends on small noise levels~\cite{karras2022elucidating}, large noise levels affect the diversity of generated results~\cite{song2020improved}. More precisely, the well-known diffusion model, EDM~\cite{karras2022elucidating}, targets the training and sampling effort on small noise levels. Their prioritization strategy dramatically reduces the number of sampling steps and achieves compelling visual quality, especially for rendering image details, \eg, human hair curls and skin pores. 

To investigate the performance of diffusion-based models in synthesizing plausible structures, we evaluate several cutting-edge diffusion models on BIKED~\cite{regenwetter2021biked}, an open-source dataset of 2D bicycle structures. With the exception of Denoising Diffusion Probabilistic Models (DDPM)~\cite{ho2020ddpm}, which utilize tremendous sampling steps and thus excel in structure modeling, other cutting-edge diffusion models~\cite{song2022ddim, karras2022elucidating} tend to generate a large portion of designs that are not plausible. We also notice that most of the commonly used metrics, \eg, FID, Learned Perceptual Image Patch Similarity (LPIPS)~\cite{zhang2018lpips} and Structural Similarity Index (SSIM)~\cite{SSIM}, do not align well with human evaluation of the plausibility of the results. In particular, the EDM~\cite{karras2022elucidating} generates images of excellent visual quality, with a FID of $7.84$, however, performs the worst regarding to plausibility, \ie, 16.6\% of the generated bicycles are not plausible. We assume that the EDM's prioritization of visual quality compromises the plausibility of generated images. 
\begin{figure}[t]
\fontsize{8}{10}\selectfont
  \centering
     \centering
     \includegraphics[width=\linewidth]{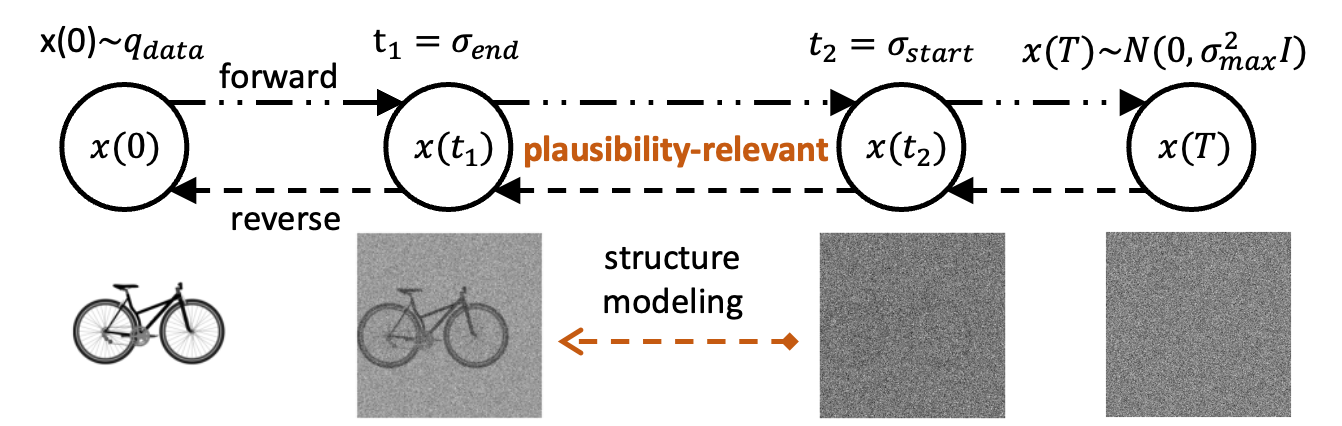}
    \caption{Forward and reverse processes of diffusion models in directed graphs, with a highlight on the plausibility-relevant range of noise levels, \ie, $[\sigma_{\text{end}},\sigma_{\text{start}}]$.}\label{fig:figure_encoding_decoding}
  \end{figure}
  
Rather than targeting a better FID value as the ultimate goal, our work aims to employ diffusion-based generative models to generate high-plausibility designs. Therefore, in evaluating generative models our work leverages three performance metrics:
\begin{itemize}
    \item Design Plausibility Score (DPS), assessed by a human evaluator based on a predefined standard. DPS takes values from one to five (from the least plausible design to the most plausible one).
    \item Plausible Design Rate (PDR), that is the proportion of plausible designs (DPS$=5$) for $1\,000$ evaluated images.
    \item Fr\'echet Inception Distance.
\end{itemize} 

\begin{figure}[t]  
  \centering
   \includegraphics[width=\linewidth, trim=3mm 2mm 7mm 2mm, clip]{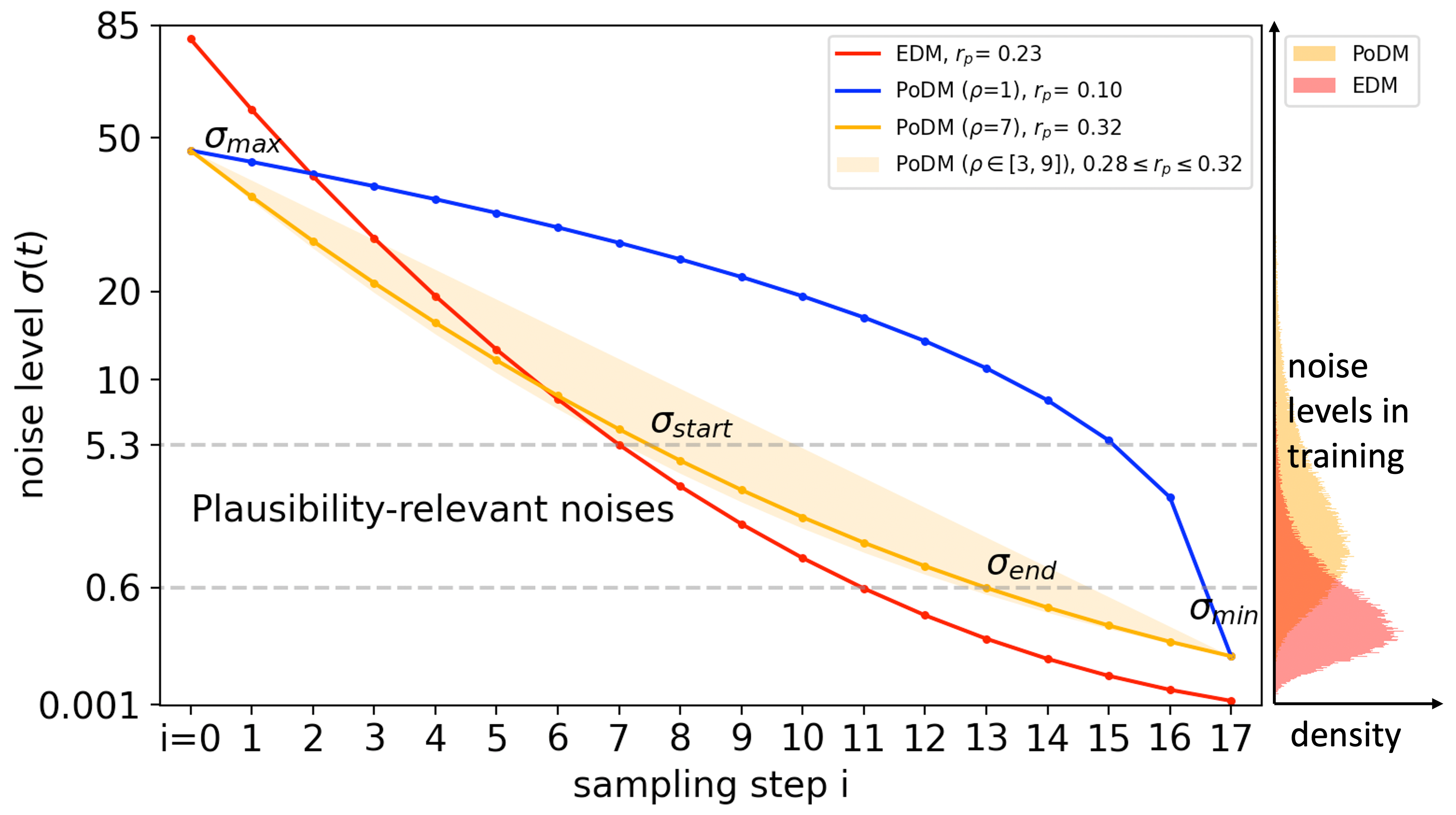}
    \caption{Comparison of noise schedules of our Plausibility-oriented Diffusion Model (PoDM) and the well-known diffusion model (EDM). PoDM intentionally focuses on noise levels within the plausibility-relevant range during both training and sampling. }
    \label{fig:noise_schedule}
\end{figure}

We assume that there is a range of noise levels that are crucial to structure modeling, and focusing on this range of noise levels can alleviate the issue of poor plausibility in generated designs. We illustrate this assumption with the diffusion-based modeling process in~\cref{fig:figure_encoding_decoding}. To determine this plausibility-relevant range of noise levels, we simulate the diffusion process on real structural designs and trace the distribution of pixel values as the noise level increases. We observe that the disappearance of the structural signal has a clear corresponding phase in the development of pixel-value distributions. Following this observation, we think that in the sampling process, the noise levels within this phase play a crucial role in the final structure. We design two techniques to statistically determine this range of noise levels. To evaluate the efficiency of prioritizing the determined range of noise levels, we modify the training and sampling procedures of EDMs, and we display our noise schedules in~\cref{fig:noise_schedule}. Compared to the default EDM, our modification increases the PDR from $83.4\%$ to $\textbf{93.5\%}$, which is almost comparable to DDPM, with a PDR of $94\%$, but requires only $\textbf{1/15}$ of the DDPM's sampling time. Additionally, this prioritizing strategy brings slight improvement in FID from $7.84$, achieved by EDMs, to $\textbf{4.87}$. This compelling result indicates the existence of the plausibility-relevant noise levels and the effect of our prioritization strategy. We collectively refer our implementation as the Plausibility-oriented Diffusion Model (PoDM), which leverages noise scheduling to achieve high plausibility in design generation. 

Given that our PoDM has accurately modeled the structural design space, we further test its performance in incorporation with modern image-editing methods, \eg, inpainting, interpolation via latent space and point-based dragging, and hereby manipulating structure.

Finally, we conclude our contributions as follows:
\begin{itemize}
    \item We observe the range of noise levels that are relevant to the plausibility of outcomes and design two techniques to statistically determine it for a given dataset.
    \item We propose a new noise scheduling method for diffusion models to prioritize the target range of noise levels, hereby improving the model performance in generating plausible designs. 
    \item We employ cutting-edge image editing tools and enable semantically manipulation of complex structural designs.
\end{itemize}

\section{Related work}\label{sec:related-work}

\paragraph{DGM-driven structure generation}
The field of Deep Generative Models (DGMs) is developing with an astonishing speed, as some models can generate novel data that is able to fool humans, especially in image synthesis~\cite{karras2019stylebased, istyleGAN, LoGAN, karras2022elucidating}. Novel research proposes to employ DGMs' generation power to generate structure designs and yields models with decent performance, \eg, PaDGAN~\cite{wei2020padgan} and BézierGAN~\cite{wei2018BezierGAN} for the UIUC Airfoil shapes~\cite{UIUCAirfoilDB}; and Self-Attention Adversarial Latent Autoencoder (SA-ALAE)~\cite{fan2023adversarial} for complex designs of automotive parts. Meanwhile, Regenwetter~\etal~\cite{regenwetter2021biked} introduces the BIKED dataset to challenge DGMs in generating complex structural designs, which is a suitable dataset for exploring the performance of DGMs in structure generation. However, to the best of our knowledge, there have not been any convincing result in generating BIKED images. The only known result to us is achieved by~\cite{regenwetter2021biked} with Variational Autoencoders (VAEs)~\cite{kingma2022autoencoding}, but the generated designs are barely recognizable. 

\paragraph{Evaluation of generative models}
Evaluating generated images is a challenge because the model is tasked to generate new images and thus there is no ground truth. There are metrics based on statistical distance between source image set and generated image set, \eg, the Earth-Mover Distance (EMD), the Kullback-Leibler divergence (KL-divergence), and also metrics for measuring perceptual distance with numerical method, \eg, Structural Similarity Index Measure (SSIM)~\cite{SSIM} and Multi-Scale Structural Similarity Index Measure (MS-SSIM)~\cite{MS-SSIM}. Several metrics are motivated by deep networks, \eg, Inception Score (IS)~\cite{salimans2016improved}, Fr\'echet Inception Distance (FID)~\cite{heusel2018TTUR} and Kernel Inception Distance (KID)~\cite{betzalel2022study} for visual quality, and Learned Perceptual Image Patch Similarity (LPIPS)~\cite{zhang2018lpips} for perceptual similarity, which are claimed to agree better with human judgments than metrics that are not driven by networks. 

\paragraph{Diffusion-based generative modeling}
Diffusion models (DM)~\cite{sohldickstein2015deep} present a novel idea of capturing data distribution and image generation. DMs did not attract much attention until the convincing implementation of Denoising Diffusion Probabilistic Model (DDPM)~\cite{ho2020ddpm}, which leverages tremendous sampling time to generate images with quality comparable to GANs. A further developed diffusion model, Denoising Diffusion Implicit Model (DDIM)~\cite{song2022ddim}, improves generation speed by trading-off image quality. Meanwhile, in the domain of score-based generation, Song~\etal~\cite{song2021scorebased} propose to use a Stochastic Differential Equation (SDE) for the forward process and a corresponding reverse-time SDE for sampling, which allows continuous diffusion processes. SDE derives a deterministic sampling process based on a corresponding ordinary differential equation (ODE), that enables identifiable encoding-decoding and more importantly flexible data manipulation via latent space. Then, Karras~\etal~\cite{karras2022elucidating} clean the design space of diffusion-based generative models and propose a novel framework, denoted as EDM. EDM achieves a new state-of-the-art performance on generation of CIFAR-10~\cite{krizhevsky2010cifar} and ImageNet-64~\cite{DenDon09Imagenet}.

Diffusion-based generative models have been introduced in controlling generation and data manipulation, \eg, interpolation via latent space~\cite{ho2020ddpm, song2021scorebased, song2022ddim, dhariwal2021diffusion}, free-form inpainting~\cite{song2021scorebased,lugmayr2022repaint} and point-based dragging~\cite{shi2023dragdiffusion}. These image editing methods tend to be applied on natural images, \eg, CelebA~\cite{huang2018celeba}, LSUN bedroom images~\cite{yu2016lsun} and ImageNet~\cite{DenDon09Imagenet}.

\section{Plausibility-oriented Diffusion Modeling}
\label{sec:PoDM}
In perturbation experiments with increasing noise levels on BIKED~\cite{regenwetter2021biked}, as the results shown in~\cref{sec:noise_scope}, we observe that the structural design fades away over a range of noise levels, and this range is clearly identifiable in the development of the pixel-value distribution. We think that this range is the plausibility-relevant range of noise levels. Here, we propose two techniques to help with determining this range statistically. In~\cref{sec:training_sampling_PoDM}, we present novel training and sampling procedures that target their efforts to the determined range of noise levels.

Although our observations and determinations are based solely on the BIKED dataset, they are equally valid to other structural designs. In fact, the BIKED images are created from CAD models, which are the standard way of developing structures today.

\subsection{Background}
\label{background}
We built up our contribution based on the stochastic differential equation (SDE) model of the diffusion process~\cite{song2021scorebased,karras2022elucidating}. Given a data point $\vx\in\R^d$, we corrupt it with the following forward It\^o SDE~\cite{oksendal2013stochastic}:
\begin{equation}
\label{eq:forward-SDE}
\ud\vx = f(\vx, t) \ud t + g(t)\ud\vB,
\end{equation}
where $f\colon \R^d \times [0, T] \rightarrow \R^d$ is the drift vector, $g\colon [0, T] \rightarrow \R$ is the dispersion coefficient, and $\vB\in \R^d$ is the standard Brownian motion. Notably, $f$ and $g$, are pre-determined by the user and have no trainable parameters. The corresponding reverse-time SDE is~\cite{anderson1982reverse}:
\begin{equation}
\label{eq:reverse_sde}
\ud \vx = [f(\vx, \tau) - g(\tau)^2 \nabla_{\vx} \log p(\vx, \tau) ] \ud \tau + g(\tau)\ud \vec{B}_{\tau},
\end{equation}
where $\tau$ goes from $T$ to 0, $p(\vx, \tau)$ is the probability density of $\vx$ at $\tau$ in the forward process, and $\nabla_{\vx} \log p(\vx, \tau)$ is known as the score function. The flow of probability mass in~\cref{eq:reverse_sde} can be equivalently described by an ordinary differential equation (ODE)~\cite{MaoutsaRO20,song2021scorebased,karras2022elucidating}:
\begin{equation}
\label{eq:ode}
\frac{\ud \vx}{\ud \tau} = f(\vx, \tau) - g(\tau)^2 \nabla_{\vx} \log p(\vx, \tau).
\end{equation}
We follow the choice of the drift and dispersion terms in~\cite{karras2022elucidating} (a.k.a.~EDM): 
$$
f(\vx,\tau) =0, \quad g(\tau) =\sqrt{2\sigma(\tau)}, \quad \sigma(\tau) = \tau,
$$
and the score function is approximated by $\nabla_{\vx} \log{p(\vx, \tau)} = (D_{\theta}(\vx;\sigma(\tau)) - \vx) /{\sigma(\tau)}^2$, where $D_{\theta}$ is a neural network trained on samples drawn from the forward SDE (see~\cite{karras2022elucidating} for details on the loss function). Due to the above choice, $\sigma$ and $\tau$ are interchangeable henceforth. To solve/sample from~\cref{eq:ode}, an $N$ time-step discretization is used with the following noise schedule: $\sigma_{N} = 0, \forall i\in [0..N-1]$:
\begin{equation}
\label{eq:noise_sampling}
\sigma_i = 
    {\left( \sigma_{\text{max}}^{\frac{1}{\rho}} + \frac{i}{N-1} (\sigma_{\text{min}}^{\frac{1}{\rho}} - \sigma_{\text{max}}^{\frac{1}{\rho}}) \right)}^\rho,
\end{equation}
where $\sigma_0 = \sigma_{\text{max}}$ and $\sigma_{N-1} = \sigma_{\text{min}}$. EMD recommends the setting: $\sigma_{\text{min}} = 0.002,\sigma_{\text{max}} = 80, \rho=7$. 
In the stochastic sampling procedure, we denote by $\vx_i$ the data point obtained at $\sigma_i$. We first increase the noise level slightly and perturb $\vx_i$:
\begin{align}
\label{eq:churn}
\!\!\!\!\!\!\vx_i^\prime &= \vx_i + \sqrt{\hat{\sigma}_i^2 - {\sigma}_i^2}\mathcal{N}(\mathbf{0}, S_{\text{noise}}^2\mathbf{I})\\
\!\!\!\!\!\!\hat{\sigma}_i &= \sigma_i\left(1 + \mathds{1}_{[S_{\text{min}},\: S_{\text{max}}]}(\sigma_i)\min\left(S_{\text{churn}}/N, \sqrt{2}-1\right)\right)
\end{align}
where $S_{\text{churn}}$ controls the degree of randomness in sampling: $S_{\text{churn}} = 0$ realizes deterministic generation. Afterwards, we apply the reverse-time ODE (~\cref{eq:ode}) with $\vx_i^\prime$ from $\hat{\sigma}_i$ to $\sigma_{i+1}$. The default settings of stochastic sampling are: $S_{\text{churn}} = 40,\: S_{\text{min}} = 0.05,\: S_{\text{min}} = 50,\: S_{\text{noise}} = 1.003$.
The training data of $D_{\theta}(\vx;\sigma(t))$ are sampled from ~\cref{eq:forward-SDE} with a log-normal distribution:
\begin{equation}
\label{eq:noise_training}
\ln(\sigma) \sim \mathcal{N}(P_{\text{mean}},\: P_{\text{std}}^2).
\end{equation}
In~\cite{karras2022elucidating}, the following empirical setting 
is suggested: $P_{\text{mean}} = -1.2, P_{\text{std}} = 1.2$.

\begin{figure*}[t]
    \centering
    \begin{subfigure}[b]{\textwidth}
    \includegraphics[width=\textwidth]{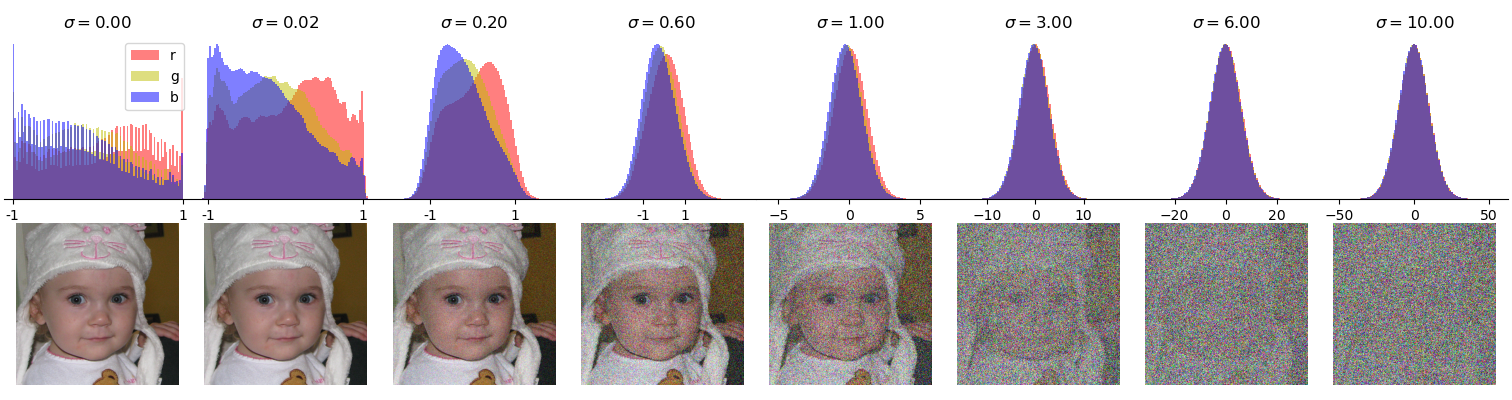}
    \caption{FFHQ~\cite{karras2019stylebased} (resolution: $64\times 64$).}
    \label{fig:FFHQ_analysis}
    \end{subfigure}\\
    \begin{subfigure}[b]{\textwidth}
    \includegraphics[width=\textwidth]{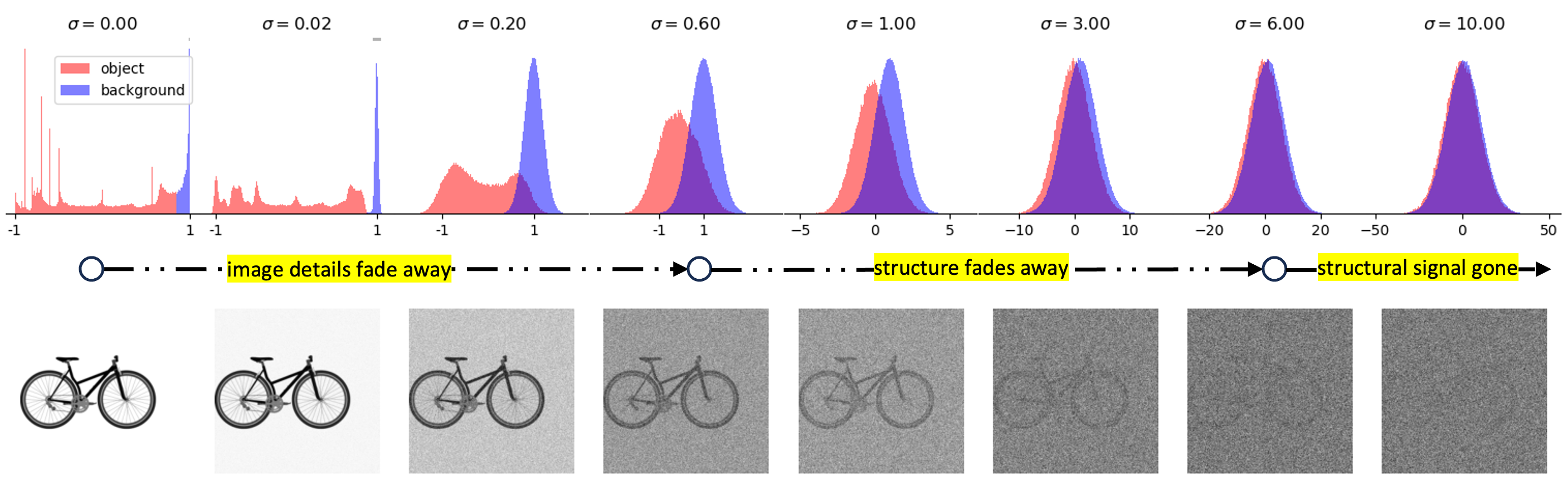}
    \caption{BIKED~\cite{regenwetter2021biked} (resolution: $256\times 256$).}
    \label{fig:biked_analysis}
    \end{subfigure}
    \caption{Pixel-value distribution in perturbation experiments. We notice that the perturbation gradually transforms the pixel values into a Gaussian distribution. With structural design images like BIKED~\cite{regenwetter2021biked}, we can determine the stages by first separating the object and background pixels and then tracking the development of the pixel-value distribution. }
    \label{fig:image_analysis_plot}
\end{figure*}
\begin{figure}
    \centering
    \begin{subfigure}[t]{0.23\textwidth}
        \includegraphics[width=\textwidth]{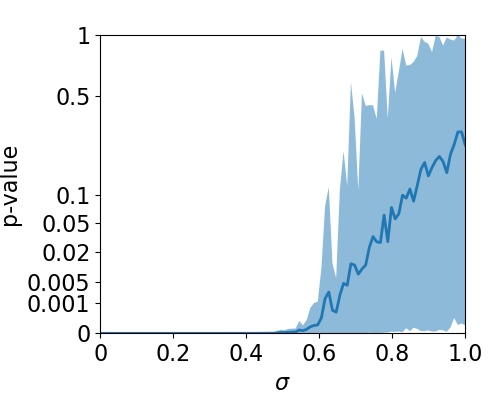}
        \caption{Shapiro-Wilk test. To determine the noise level at which object pixel values are Gaussian distributed.}
        \label{fig:biked-shapiro_O}
    \end{subfigure}
    \hfill
    \begin{subfigure}[t]{0.23\textwidth}
        \includegraphics[width=\textwidth]{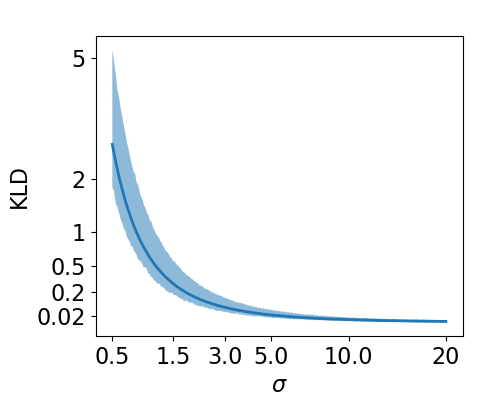}
        \caption{Kullback-Leibler Divergence. To determine the noise level at which object pixels are indistinguishable from background pixels.}
        \label{fig:biked-KLD_BO}
    \end{subfigure}
    \caption{Statistical analysis on BIKED. Shaded regions represent the measured values over $100$ random source samples, plotting their mean values in a blue curve.}
    \label{fig:statistical_analysis_biked}
\end{figure}

\subsection{Noise level relevant to plausibility}
\label{sec:noise_scope}
We conjecture that in the forward/backward process, the pixel-value distributions of a structural design are substantially different from that of a real-world image. We validate it by running the forward process with fine-grained noise levels: starting with a source image $\vx(0)$ (the pixel values are standardized to $[-1,1]$ before adding the Gaussian noise), we keep adding small Gaussian noises thereto until its pixel-value distribution becomes indistinguishable from a Gaussian:
\begin{equation}
\vx(t) = \vx(0) + \sigma(t)\mathcal{N}(\mathbf{0},\mathbf{I}), \quad \sigma(t) = 0.1t.
\end{equation}
We show the pixel-value distribution at intermediate time steps in~\cref{fig:image_analysis_plot} for $100$ randomly selected images respectively from FFHQ~\cite{karras2019stylebased} and BIKED~\cite{regenwetter2021biked} images. For the BIKED images, which consist of a unique object and a monotone background, we depict the pixel-value distribution separately for the object and the background (the red and blue curves). We observe that (1) the distribution of the bicycle object is significantly distinguishable from that of the background while object pixel-value distribution is converging to a Gaussian, \ie, for $\sigma\in [0,0.6]$; (2) the bicycle structure fades away while the object pixel-value distribution overlaps more with that of the background, \ie, for $\sigma\in[0.6, 6.0]$; (3) the signal of bicycle structure is gone, \ie, for $\sigma>6$. For FFHQ images, since multiple objects are present therein and there is no monotone background, we track the distribution for each RGB channel. In contrast, the pixel-value distribution of the three channels mostly overlaps with each other, and therefore the forward process does not exhibit discernible stages in which the objects are destroyed by the noise. 

Empirically, for structural images, we notice that the noise range ($\sigma\in[0.6, 6]$ in~\cref{fig:image_analysis_plot}) in which the bicycle structure fades away determine the plausibility of the generation. We denote this noise range as $[\sigma_{\text{end}},\sigma_{\text{start}}]$ and propose two techniques to determine this interval in general.
\textbf{Technique 1:} \textit{Choose $\sigma_{\text{end}}$ to be the largest noise level at which the object pixel values are not normally distributed according to the Shapiro-Wilk test with a significance level of $0.01$.}

In a sampling process, $\sigma_{\text{end}}$ is the noise level at which the generation is structurally finalized and object pixel values remain normally distributed. Denoising with noise levels of $\sigma\leqslant\sigma_{\text{end}}$ performs a refinement, during which the backward process approximates the object pixel values from a normal distribution to a local data distribution.
To estimate $\sigma_{\text{end}}$, we propose to use the Shapiro-Wilk test~\cite{shapiro_analysis_1965} to track the distribution of the object's pixel values during the perturbation test. As displayed in~\cref{fig:biked-shapiro_O}, the measured $p$-value increases with the noise level. 

\textbf{Technique 2:} \textit{Choose $\sigma_{\text{start}}$ to be the noise level at which pixel-value distributions of object and background are sufficiently close, \ie, when the Kullback–Leibler divergence (KL-divergence) reduces to $0.02$.}

In the synthesis of structural design images, $\sigma_{\text{start}}$ is the noise level at which the structural formation begins, \ie, pixels of objects begin to distinguish themselves from pixels of the background. To measure such a difference, we first approximate the pixel-value distributions of the object and background with Gaussians, respectively, and then compute the KL-divergence between the two Gaussian approximations, \ie,
\begin{equation}
\log\left(\frac{\hat{s}_{\text{b}}}{\hat{s}_{\text{o}}}\right)+ \frac{(\hat{s}_{\text{o}})^2 + (\hat{m}_{\text{o}}-\hat{m}_{\text{b}})^2}{2\hat{s}_{\text{b}}^2},
\end{equation}
where $(\hat{m}_{\text{o}}, \hat{s}_{\text{o}})$ and $(\hat{m}_{\text{b}}, \hat{s}_{\text{b}})$ are the mean and standard deviation of the pixel values estimated from the object and background, respectively. As shown in~\cref{fig:biked-KLD_BO}, $\sigma_{\text{start}}$ is taken when KL-divergence converges.

Our work gives an insight into defining the plausibility-relevant range of noise levels so that the training and sampling effort can prioritize this range. By implementing our two techniques, for BIKED images, we determine $\sigma_{\text{end}}$ to be $0.6$ and $\sigma_{\text{start}}$ to be $5.3$. In practice, if computational power allows, we suggest slightly extending the range $[\sigma_{\text{end}},\sigma_{\text{start}}]$ to make better usage of the model capacity.

\subsection{Plausibility-oriented training and sampling procedures}
\label{sec:training_sampling_PoDM}
We propose crucial modifications to the noise schedules in both the training and the sampling procedures, which affects significantly the plausibility of the generated images. 

\paragraph{Training noise density} 
For the structure images, the generation of the structure takes place mostly in the noise range $[\sigma_{\text{end}}, \sigma_{\text{start}}]$ while the noise levels that are too small or large have marginal effects on the plausibility of the final outcome. Hence, it is sensible to sample more noise levels in this interval from the forward SDE. We modify the sample distribution in~\cref{eq:noise_training} based on $[\sigma_{\text{end}}, \sigma_{\text{start}}]$:
\begin{align}
&\ln(\sigma) \sim \mathcal{N}(\mu,\: \zeta^2) \label{eq:noise_training_PoDM}, \\
& \mu = \frac{1}{2}\left(\ln(\sigma_{\text{start}}) + \ln(\sigma_{\text{end}})\right) \label{eq:calculating_mean}, \\
&\zeta = \frac{1}{2}\left(\ln(\sigma_{\text{start}}) - \ln(\sigma_{\text{end}})\right)\label{eq:calculating_std},
\end{align}
which implies $\Pr(\sigma \in [\sigma_{\text{end}}, \sigma_{\text{start}}]) \approx 68\%$. In this method, the majority of the noise levels are drawn in $[\sigma_{\text{end}}, \sigma_{\text{start}}]$ while we have ca. $32\%$ probability to sample noise levels at the beginning and the end of the forward process. 


\paragraph{Sampling noise schedule} 
In the sampling (\cref{eq:ode}), there are two important factors w.r.t.~the noise levels: (1) the noise range $[\sigma_{\text{min}}, \sigma_{\text{max}}]$ in which we apply the ODE (\cref{eq:ode}) and (2) the decaying noise schedule. For the former, we determine the range based on the training noise density as follows: \cref{eq:noise_training_PoDM} implies that the score function $\nabla_{\vx} \log p(\vx, \sigma)$ is trained on noise levels drawn almost in $[\mu-3\zeta,\mu+3\zeta]$, \ie, $\Pr(\log(\sigma) \in [\mu-3\zeta,\mu+3\zeta]) \approx 99.7\%$. Therefore, when sampling new images, applying the reverse-time ODE out of $[\mu-3\zeta,\mu+3\zeta]$ requires the score function to extrapolate, which we have no guarantee about its accuracy. Hence, we set 
\begin{align}
\log \sigma_{\text{min}} &= \mu -3\zeta = 2\log\sigma_{\text{end}} - \log\sigma_{\text{start}}\;, \\
\log \sigma_{\text{max}} &= \mu +3\zeta = 2\log\sigma_{\text{start}} - \log\sigma_{\text{end}}\; .
\end{align}
For the latter, we follow the exponential decay in~\cref{eq:noise_sampling}, where, in addition, we tune the hyperparameter $\rho$ for the BIKED dataset. In~\cref{tab:fine-tuning_exponent}, we summarize the tuning results from a simple grid search, where $\rho=7$ is the best setting. Also, we observe that the performance metrics (\eg, FID, DPS, and PDR) are quite sensitive to $\rho$, suggesting that tuning this parameter is necessary across different structural image data. Moreover, we calculate the proportion of the noise levels $\{\sigma_{N-1}, \ldots, \sigma_{0}\}$ (determined by \cref{eq:noise_sampling}) falling into the plausibility-relevant range $[\sigma_{\text{end}},\sigma_{\text{start}}]$, which we call the prioritization density $r_p$. It measures how much training effort is targeted at the structure modeling. In~\cref{fig:noise_schedule} and~\cref{tab:fine-tuning_exponent}, we show $r_p$ with varying hyperparameter $\rho$, and we observe that the performance metrics are positively related to it.

\section{Evaluation and results}
\label{evaluation_results}
\begin{figure}[t]
    \centering
    \includegraphics[width=\linewidth]{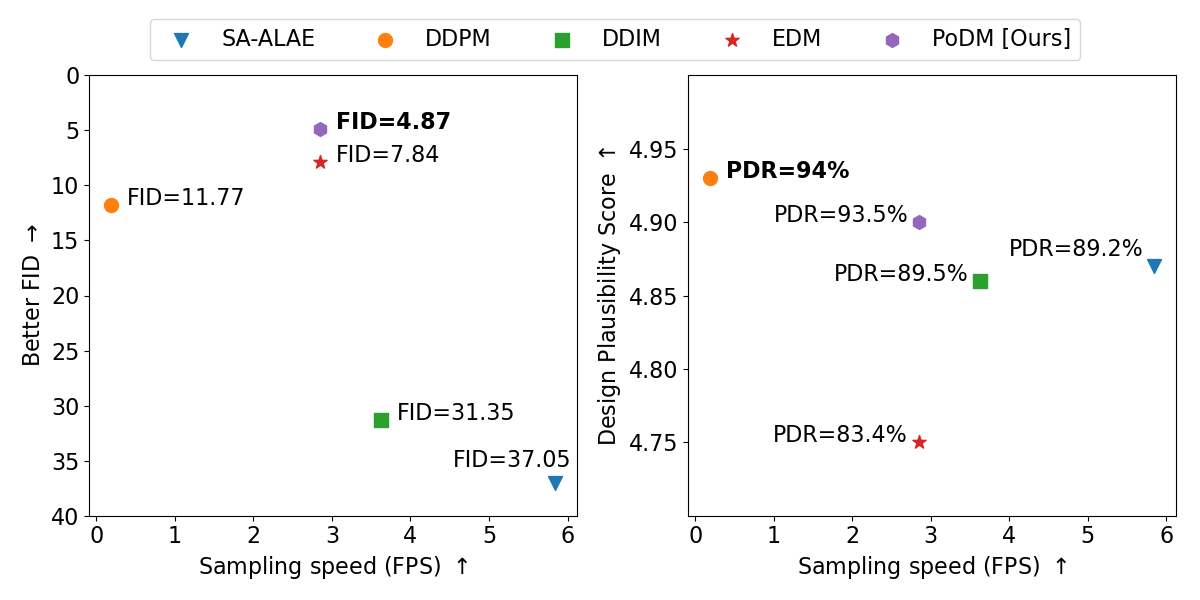}
    \caption{Comprehensive overview. We plot the results based on sampling speed, taking into account that the longer the sampling time is, the better the results are. EDM~\cite{karras2022elucidating} excels in visual quality, but it seems that EDM heavily compromises the image plausibility. Our PoDM outperforms the EDM in visual quality and plausibility with the same sampling speed; and the plausibility of our generated design is comparable to that of DDPM~\cite{ho2020ddpm}, but with a significantly faster sampling speed.}
    \label{modelEvaluationResults}
\end{figure}

\begin{table}[t]
\fontsize{8}{10}\selectfont
    \centering
    \begin{tabular}{*{6}{c}} 
    \toprule
    $\rho $ & $r_p$ & FID$\downarrow$ & DPS$\uparrow$ & PDR$\uparrow$\\
    \midrule
    $1$ & $0.10$ & $12.67$ & $4.70$ & $82.8\%$\\
    $3$ & $0.28$ & $5.64$ & $4.81$ & $88.6\%$\\
    $5$ & $0.31$ & $5.25$ & $4.88$ & $89.6\%$\\
    $7$ & $0.32$ & $\textbf{4.87}$ & $\textbf{4.90}$ & $\textbf{93.5\%}$\\
    $9$ & $0.32$ & $5.18$ & $4.87$ & $91.5\%$\\
    \bottomrule
    \end{tabular}
    \caption{Exponent $\rho$ tuning for PoDM. $\rho$ is the exponent in creating sampling schedules; $r_p$ represents the density of noise levels within the prioritizing range during sampling. Notably, PDR and DPS are positively related to $r_p$.}
    \label{tab:fine-tuning_exponent}
\end{table}

\subsection{Training configurations}
Our work utilizes the model architecture from the DDIM~\cite{song2022ddim} repository for all diffusion-based models, which follows the U-Net proposed by Ho \etal~\cite{ho2020ddpm}. More precisely, the implemented model has six feature map resolutions from $256 \times 256$ to $4 \times 4$, one residual block for each upsampling/downsampling and an attention layer at the feature map resolution of $16 \times 16$. For sampling with DDPM and DDIM, we use the same trained model with default training settings, \ie, timesteps of $1\,000$ and linear schedule of $\beta$ with $\beta_0 = 1e-4, \beta_T = 0.02$. For EDM, we remove EDM's preconditions, since they did not bring much enhancement to the results according to their experiments, and implement their noise schedules for both training and sampling with default parameters, \ie, $\sigma_{\text{min}} = 2e-3, \sigma_{\text{max}} = 80, \rho = 7, P_{\text{mean}} = -1.2, P_{\text{std}} = 1.2$. For our PoDM, we determine $\sigma_{\text{start}} = 5.3$ and $\sigma_{\text{end}} = 0.6$ by analyzing the BIKED dataset and inherit the loss function from EDM. For stochastic sampling in both EDM and our PoDM, we allow the ``churn'' modification (\cref{eq:churn}) for all sampling steps, \ie, $S_{\text{min}}=0,\:S_{\text{max}}=\infty$, and set the $S_{\text{churn}}$ to $5$. For DDIM, we use $50$ as the number of sampling steps, whereas for both EDM and our PoDM, the number of sampling steps is set to $18$. The set ``Standardized Images'' from BIKED Dataset~\cite{regenwetter2021biked} contains $4\,512$ grayscale pixel-based images with original shape of $1\,536\times710$. We pad them with background pixels to a square form with the shape of $1\,536\times 1\,536$. Then we reshape these images into a resolution of $256\times256$ with scale of $[-1, 1]$, in order to maintain the height-width ratio and ease the complexity in generation. From the whole dataset, we randomly select 100 images for validation, $1\,000$ images for testing and the rest images for training. We run training on four NVIDIA DGX-2's Tesla V100 GPUs with batch size of $32$ and learning rate of $5e-5$. Model parameters are saved every $1\,000$ steps. If the loss converges, we keep training until $100\,000$ steps and then stop it when the denoising loss does not decrease for $20$ epochs. For each model, we select the best-performing model within the saved checkpoints in the last $20\,000$ steps.

In this section, we compare our Plausibility-oriented Diffusion Model (PoDM) with other cutting-edge diffusion-based models, \ie, vanilla DDPM~\cite{ho2020ddpm}, DDIM~\cite{song2022ddim} and EDM~\cite{karras2022elucidating}. We also include SA-ALAE~\cite{fan2023adversarial} in the comparison, considering that SA-ALAE was recently proposed to address the generation of complex structural designs.

\subsection{Evaluation procedures}
\label{sec:metrics}
In our work, we evaluate the generative models in terms of sampling speed, visual quality and design plausibility. For the sampling speed, we simply record the sampling time for generating $5\,000$ images and calculate the sampling speed in FPS (frames per second) for each model. For visual quality, we further use the $5\,000$ images generated and calculate the FID~\cite{heusel2018TTUR} between the test images and the generated images. The measured FIDs are displayed in~\cref{modelEvaluationResults}.
\begin{figure}[t]
    \centering
    \includegraphics[width=0.95\linewidth]{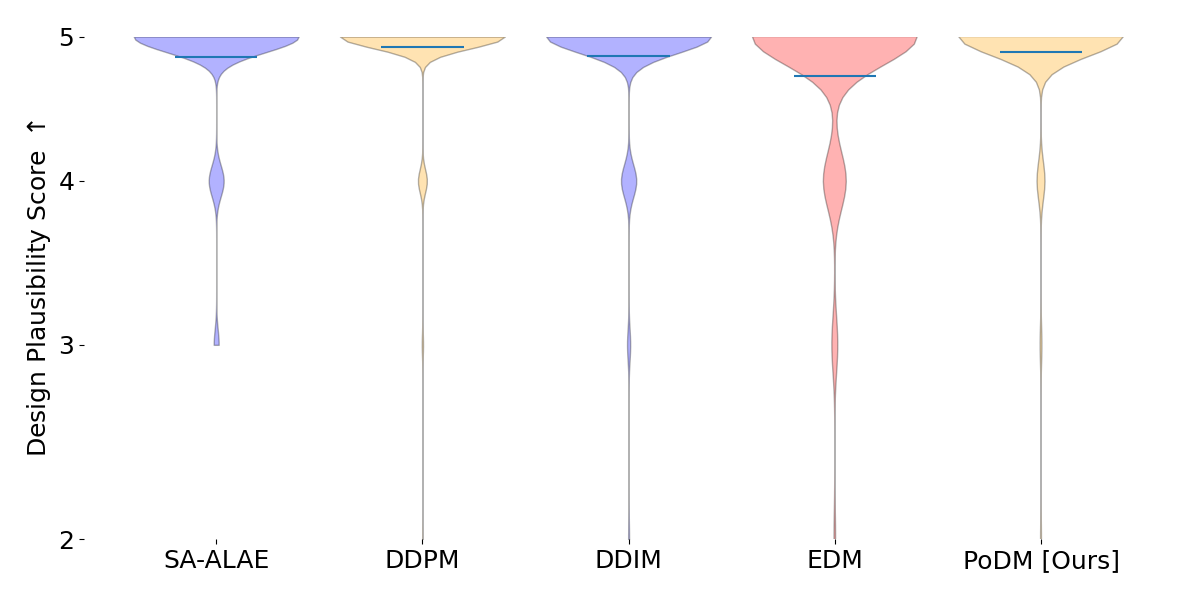}
    \caption{Overview of measured design plausibility scores. EDM~\cite{karras2022elucidating} performs poorly in design plausibility, whereas our PoDM has a comparable performance to the DDPM~\cite{ho2020ddpm}.}
    \label{fig:violinplot_model_DPS}
\end{figure}
To quantitatively evaluate the plausibility of generated designs, we implement a human evaluation method, in which the human evaluator bypasses visual qualities (\eg, blurriness and background noise) and scores the represented design in terms of plausibility. We refer to the evaluation score as the Design Plausibility Score (DPS). In this work, the generated bicycle designs are evaluated using a five-point scoring system based on the following criteria: 
\begin{itemize}
    \item No missing fundamental part;
    \item No floating material or extra part;
    \item Every part is complete;
    \item Parts are connected;
    \item Rational positioning.
\end{itemize}
For each target model in the models list, we randomly select $1\,000$ samples from the $5\,000$ generated bicycle images. We shuffle all selected images and keep tracking their DPS in a manner that associates each image’s score with its corresponding model. This experiment aims to prevent potential biases in the evaluation of the generated images by individual target models and to sustain a uniform evaluation standard across all images. We record the measured DPSs in~\cref{fig:violinplot_model_DPS} and an average DPS for each model in~\cref{modelEvaluationResults}. Besides, we calculate the Plausible Design Rate (PDR), which is the proportion of plausible designs, \ie, designs with DPS of $5$, in $1\,000$ generated images.  For each model, we also measure the LPIPS~\cite{zhang2018lpips} based on a trained AlexNet~\cite{krizhevsky2014one} and SSIM~\cite{SSIM}, and in \cref{fig:alignment_test}, we show the results of FID, LPIPS, SSIM and their correlation with DPS..

\subsection{Results}
\label{sec:model_performance}
\begin{figure}[t]
    \centering
    \includegraphics[width=\linewidth]{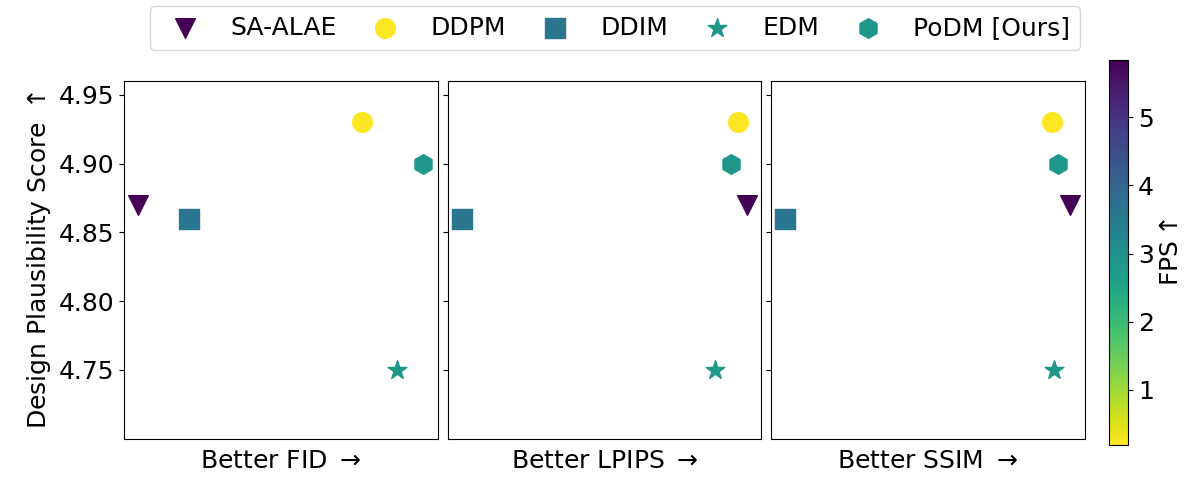}
    \caption{Alignment between human evaluation score and other metrics for generative models. These commonly used metrics do not correlate well with the human evaluation in terms of plausibility. Notably, with DDPM, DDIM and EDM there exist a trade-off among sampling speed, visual quality and design plausibility, which does not seem to be an issue with our PoDM.}
    \label{fig:alignment_test}
\end{figure}

DDPM~\cite{ho2020ddpm} requires the longest sampling time of $5.26$ seconds for each sample, but performs decently well in terms of image quality, \ie, FID of $11.77$, and design plausibility, \ie, DPS of $4.93$ with only $6.0\%$ implausible outcomes. As shown in~\cref{modelEvaluationResults}, EDM~\cite{karras2022elucidating} can significantly improve the sampling speed to $2.85$ FPS and even enhance the visual quality to a FID of $7.84$. However, EDM performs poorly in design plausibility, \ie, DPS of $4.75$ and a plausible design rate of $83.4\%$. As additional results shown in~\cref{fig:alignment_test}, DDIM and EDM demonstrate a trade-off between visual quality and plausibility of generated images, whereas the DDPM leverages extremely slow sampling speed to perform decently in both aspects. Surprisingly, our PoDM maintains the fast sampling of EDMs, achieves a compelling FID of $\textbf{4.87}$, and improves the DPS to $\textbf{4.90}$, with a remarkable plausible design rate of $\textbf{93.5\%}$.

\section{Controllable generation}
In this section, we test the PoDM's understanding of structural design space by applying cutting-edge image editing methods, \eg, interpolation via latent space, point-based dragging and inpainting, on bicycle designs.
\begin{figure*}[t]
\fontsize{8}{10}\selectfont
\centering
\begin{subfigure}[b]{\linewidth}
\centering
\includegraphics[width=\linewidth]{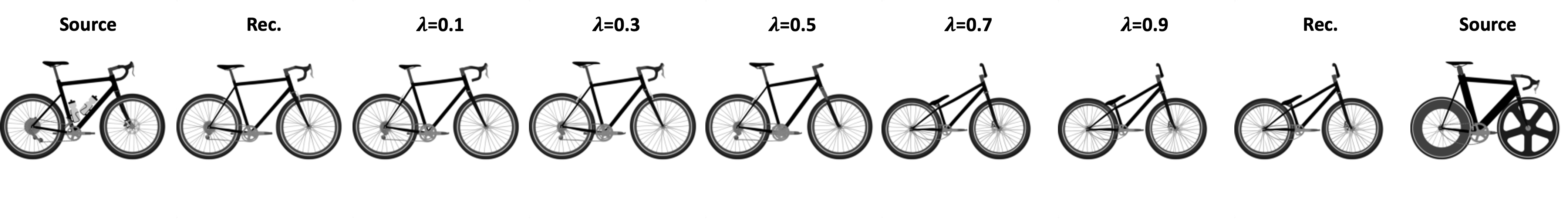}
\caption{Interpolation with latent space set at noise level $\sigma_{\text{max}}$. The reconstruction has a poor accuracy and interpolation fails to produce intermediate structures.}
\label{fig:figure_interpolation_max}
\end{subfigure}
\begin{subfigure}[b]{\linewidth}
\centering
\includegraphics[width=\linewidth]{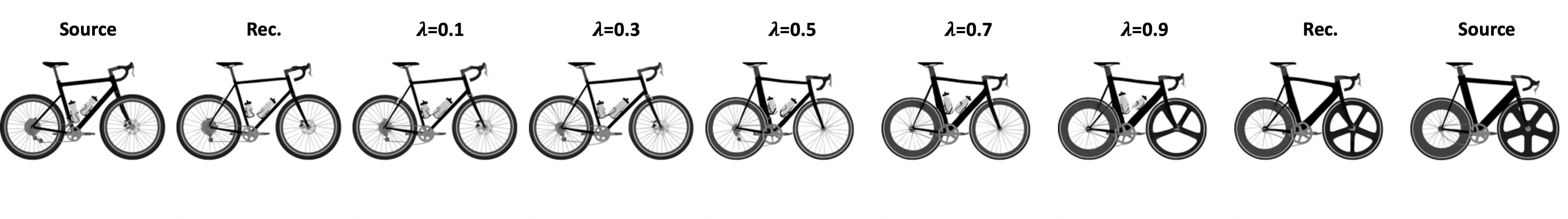}
\caption{Interpolation with latent space set at noise level $\sigma_{\text{start}}$. The interpolation displays a smooth transformation between two source structures.}
\label{fig:figure_interpolation_start}
\end{subfigure}
\caption{PoDM-driven structural interpolation via latent space set at various perturbation steps.}
\label{fig:figure_interpolation_images}
\end{figure*}
\begin{figure}[t]
    \centering
    \includegraphics[width=\linewidth]{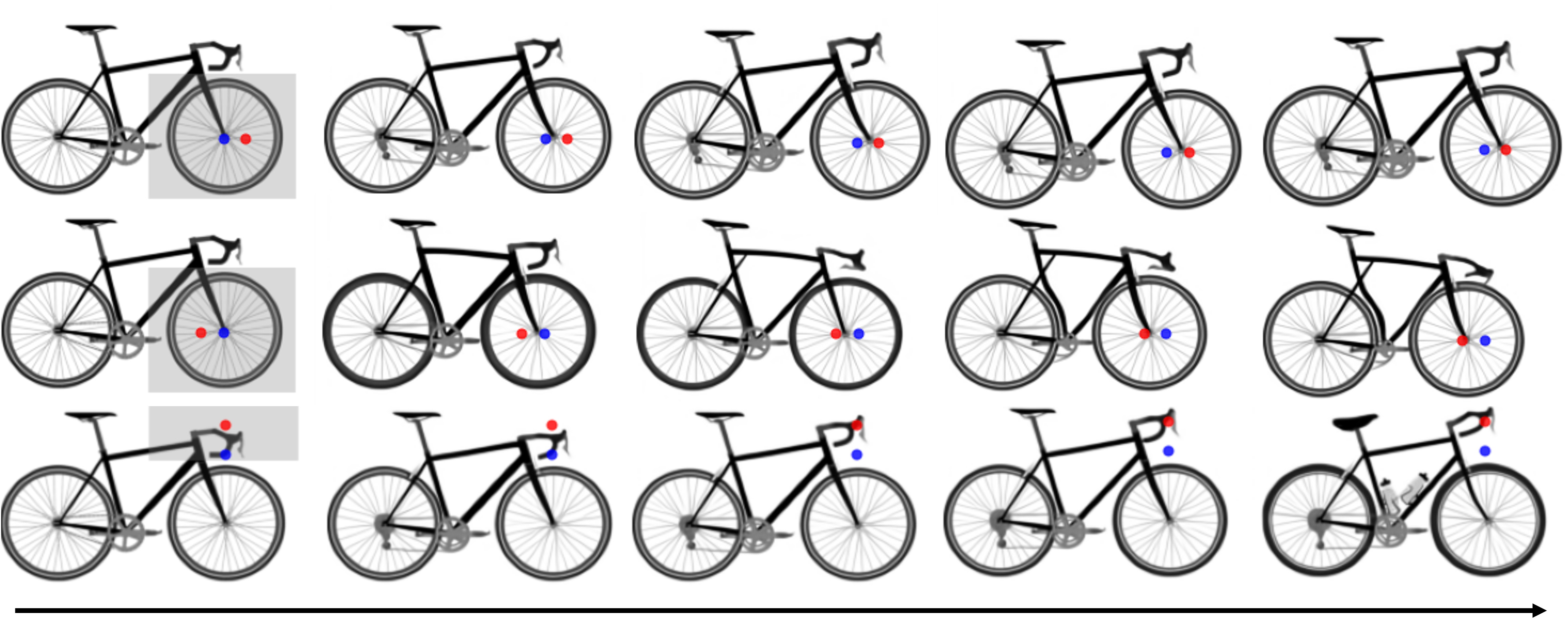}
    \caption{PoDM-driven structure editing with DragDiffusion~\cite{shi2023dragdiffusion}. From left to right, handle point is iteratively dragged from initial handle point (blue) towards the selected target point (red).}
    \label{fig:dragBIKED}
\end{figure}

\paragraph{Interpolation via latent space}
Interpolation via latent space can be quite useful in exploring structural design space. After encoding a source data $\vx(0)$ to pure noise $\vx(T)$ via the forward process, diffusion model is supposed to decode $\vx(T)$ back to $\vx(0)$ by utilizing a corresponding ODE~\cite{song2021scorebased}. However, in our implementation shown in~\cref{fig:figure_interpolation_max}, PoDM-motivated reconstruction has a poor accuracy, which might be caused by the prioritizing strategy. We argue that it is unnecessary to conduct the forward process completely, instead, perturbed images at noise level $\sigma_{\text{start}}$ retain good reversibility. Taking images at noise level $\sigma_{\text{start}}$ as latent code allows well-performing reconstruction and interpolation, as shown in~\cref{fig:figure_interpolation_start}.


\paragraph{Point-based dragging}
As a novel image editing method, point-based dragging~\cite{pan2023draggan, shi2023dragdiffusion} can precisely and iteratively ``drag" the handle point to a target point and the remaining parts of the image will be correspondingly updated to maintain the realism. We implement DragDiffussion~\cite{shi2023dragdiffusion} on BIKED images and plot the results in~\cref{fig:dragBIKED}. To our best knowledge, our work is the first to apply point-based dragging on structural design.

\begin{figure}[t]
    \centering
    \includegraphics[width=\linewidth]{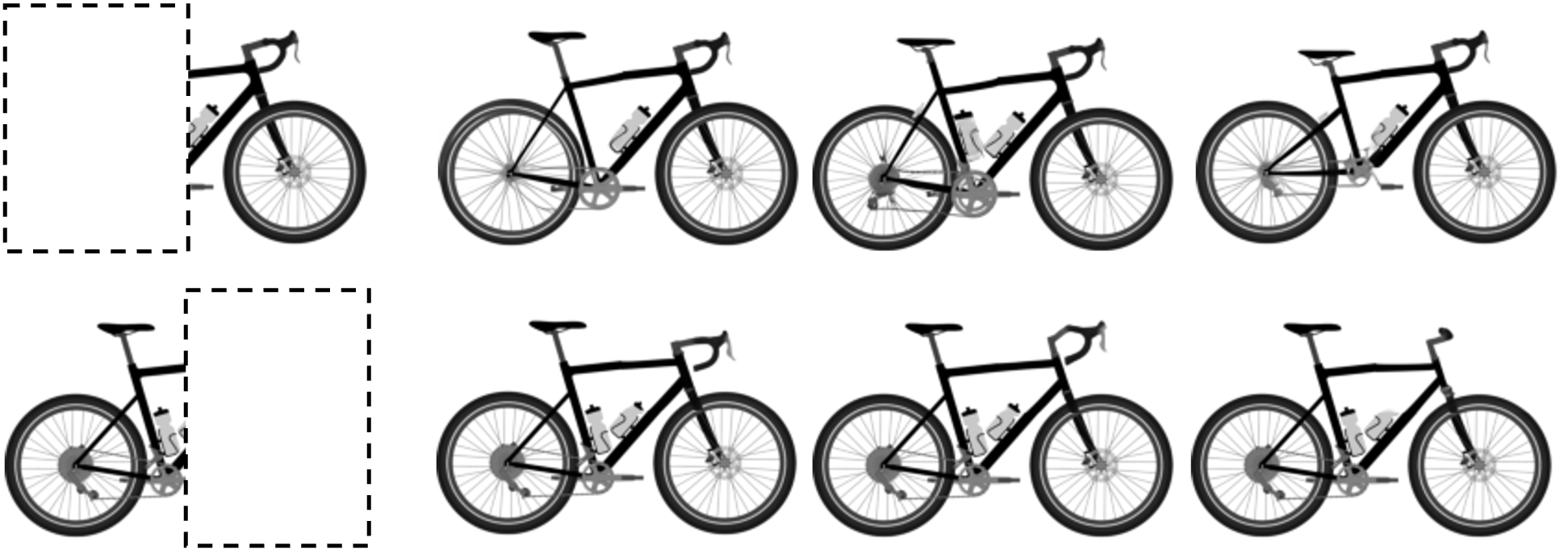}
    \caption{PoDM-driven inpainting with RePaint~\cite{lugmayr2022repaint}.}
    \label{fig:inpainting}
\end{figure}

\paragraph{Inpainting}
In an inpainting task, the generative model is tasked to generate the inpainting area to match the known part. A DDPM-based inpainting mechanism, RePaint~\cite{lugmayr2022repaint}, has achieved the state-of-the-art performance on diffusion-based inpainting tasks, by utilizing the known part as guidance at each step. We adapt RePaint to our PoDM and test it on BIKED images. The inpainting results are shown in~\cref{fig:inpainting}. 
\section{Conclusion}
We observe that the performance of the diffusion-based generative models exhibits a trade-off among visual quality, the plausibility of generated images and sampling time. We assume that there is a range of noise levels, that is responsible for the plausibility of the outcome, especially in generating structures. Following this observation, we propose a Plausibility-oriented Diffusion Model (PoDM) that leverages a novel noise schedule to prioritize this range of noise levels in both training and sampling procedures. We apply our PoDM on the well-known EDM to tackle its poor performance in generating plausible structures. Our method significantly improves the plausibility of generated images and even achieves a compelling plausibility score comparable to DDPM, but with much reduced sampling time. Additionally, we demonstrate with convincing results that the improvement in the plausibility thanks to the prioritization of the determined noise range. Further implementations of PoDM-driven image editing tools showcase PoDM's ability to semantically manipulate complex structural designs, paving the way for future work in the field of generative design. 

Our work is inspired by, but not limited to structural design generation. We believe that our observations and determinations of the stages in the diffusion process are equally applicable to images from natural scenes, and therefore beneficial for all diffusion-based synthesis tasks. In addition, we hope that our work will inspire more research on the tool for automatically evaluating plausibility of generated images and the relevance between noise level and generated features.

{
    \small
    \bibliographystyle{ieeenat_fullname}
    \bibliography{main}
}


\end{document}